\documentclass[conference]{IEEEtran}
\usepackage{cite}
\usepackage{amsmath,amssymb,amsfonts}
\usepackage{wrapfig}
\usepackage{color}
\usepackage[hyperfootnotes=false]{hyperref}
\usepackage{cleveref}
\usepackage{algorithmic}
\usepackage{graphicx}
\usepackage{textcomp}
\usepackage[dvipsnames]{xcolor}
\def\BibTeX{{\rm B\kern-.05em{\sc i\kern-.025em b}\kern-.08em
    T\kern-.1667em\lower.7ex\hbox{E}\kern-.125emX}}
\usepackage{enumitem}
\usepackage{mathtools}
\usepackage{booktabs}
\usepackage{subfigure}
\usepackage{footmisc}
\usepackage{lipsum}
\usepackage{footmisc}

\usepackage{booktabs}
\usepackage{array,multirow}

\usepackage[normalem]{ulem}

\usepackage{etoolbox}
\makeatletter
\patchcmd{\@makecaption}
  {\scshape}
  {}
  {}
  {}
\makeatletter
\patchcmd{\@makecaption}
  {\\}
  {.\ }
  {}
  {}
\makeatother

\Crefname{equation}{Eq.}{Eqs.}
\Crefname{section}{Sect.}{Sects.}
\Crefname{figure}{Fig.}{Figs.}
\Crefname{tabular}{Tab.}{Tabs.}
\Crefname{figure}{Fig.}{Figs.}
\Crefname{tabular}{Tab.}{Tabs.}

\DeclareMathOperator*{\argmax}{argmax} 

\DeclareFontFamily{OT1}{pzc}{}
\DeclareFontShape{OT1}{pzc}{m}{it}{<-> s * [1.100] pzcmi7t}{}
\DeclareMathAlphabet{\mathpzc}{OT1}{pzc}{m}{it}

\begin{document}
\title{A Neural Model for Regular Grammar Induction}

\author{
    \IEEEauthorblockN{Peter Belcak}
    \IEEEauthorblockA{
        \textit{ETH Z\"urich}\\
        Z\"urich, Switzerland \\
        belcak@ethz.ch
        }
    \and
    
    \IEEEauthorblockN{David Hofer}
    \IEEEauthorblockA{
        \textit{ETH Z\"urich}\\
        Z\"urich, Switzerland \\
        davhofer@ethz.ch
    }
    \and
    \IEEEauthorblockN{Roger Wattenhofer}
    \IEEEauthorblockA{
        \textit{ETH Z\"urich}\\
        Z\"urich, Switzerland \\
        wattenhofer@ethz.ch
    }
}

\maketitle

\begin{abstract}
Grammatical inference is a classical problem in computational learning theory and a topic of wider influence in natural language processing.
We treat grammars as a model of computation and propose a novel neural approach to induction of regular grammars from positive and negative examples.
Our model is fully explainable, its intermediate results are directly interpretable as partial parses, and it can be used to learn arbitrary regular grammars when provided with sufficient data.
We find that our method consistently attains high recall and precision scores across a range of tests of varying complexity.
\end{abstract}


\section{Introduction}
\label{section:introduction}
Finite automata, or finite state machines, are the simplest class of computational devices possessing memory.
Their memory is finite and stored exclusively as their \textit{state}, but that alone is enough to make their analysis challenging.
In contrast to simpler devices entirely characterisable by mappings of individual inputs to their corresponding outputs, finite automata may need the entire input history to determine their next state or output.

The learning of finite automata is a classical problem in computational learning theory, and can be phrased as follows: given a language $\mathcal{L}$ and a set of examples $\mathcal{E}$, use $\mathcal{E}$ to find a finite acceptor automaton $\mathcal{D}$ that accepts words in $\mathcal{L}$ and rejects words not in $\mathcal{L}$, or at least find a $\mathcal{D}$ that performs these duties with sufficient accuracy.
For instance, $\mathcal{L}$ can be the language generated by the regular expression $a^*bb^*$.
The set of examples $\mathcal{E}$ could consist of the entire $\mathcal{L}$, with us placing an additional requirement of conciseness on the learning process to arrive at a single, practical solution (say an automaton with at most three states).
Or, $\mathcal{E}$ could contain words such as $aab$ and $bbb$ marked as positive examples (the words $\mathcal{D}$ should accept), and $aaa$ and $aba$ marked as negative examples (words to be rejected by $\mathcal{D}$).

Strictly algorithmic, formal, statistical, and genetic approaches have all been proposed, but this problem remains open mainly because the solutions lack an agreed single measure of merit.
Some have argued for automata that concisely represent $\mathcal{E}$, while others preferred automata that generalised well to $\mathcal{L}$ in scenarios where $\mathcal{E}$ contains only a few examples representative of the original language.
Another degree of freedom arises in the constraints that are placed on $\mathcal{E}$.
One can insist that $\mathcal{E}$ consists of only positive examples (i.e. $\mathcal{E} \subseteq \mathcal{L}$), that $\mathcal{E}$ contains enough examples to sanction good generalisation to $\mathcal{L}$ (e.g. all repetitions of a pattern beyond a certain count signify that the pattern may be repeated indefinitely), or that $\mathcal{E}$ contains both positive and negative examples (i.e. $\mathcal{E} \cap \mathcal{L} \neq \emptyset \neq \mathcal{E} \cap \mathcal{L}^c$, where $\mathcal{L}^c$ denotes words not in $\mathcal{L}$).
We give a detailed overview of related work in \Cref{section:related_work}.

This paper proposes a neural model that learns a finite acceptor automaton for target language $\mathcal{L}$ from a given $\mathcal{E}$ consisting of positive and negative examples.
We insist that the logic of the resulting automaton is fully explainable -- fully comprehensible by humans, that the resulting automata generalise well to $\mathcal{L}$ despite being based on only $\mathcal{E}$, and that the representations are concise, or at least such that they do not contain too much redundancy of computational logic.


Internally, our model learns a regular grammar rather than a finite automaton.
Briefly, a grammar $\mathcal{G}$ is a triplet of $\mathcal{A}_T,\mathcal{A}_N,\mathcal{P}$ -- the terminal alphabet, non-terminal alphabet, and set of productions, respectively, where one of the letters of $\mathcal{A}_N$ is marked as the start symbol (the ``root'' of $\mathcal{G}$).
A left-regular grammar is a grammar in which all productions of $\mathcal{P}$ are of the form $A \to c$, $A \to \epsilon$, or $A \to Bc$ for $A,B \in \mathcal{A}_N$, $c \in \mathcal{A}_T$, $\epsilon$ the empty string.
A right-regular grammar uses productions of the form $A \to cB$ instead of $A \to Bc$.
Regular grammars are equivalent to finite automata in their language-generating power and admit straightforward conversions into each other.
We do not consider empty languages or grammars that give empty languages.

In our setting, there is a target language $\mathcal{L} = \left< \mathcal{G} \right>$ generated by the ground-truth grammar $\mathcal{G}$.
We are shown examples $\mathcal{E}$ taken from both $\mathcal{L}$ and $\mathcal{L}^c$, and our model internally learns a hypothesis grammar $\mathcal{G}'$.
In other words, we do not train neural automata to become accurate acceptors for given languages $\mathcal{L}$, but work instead with a fully explainable regular language learning parser under an acceptor training setup.
We use less information than many previous methods for grammatical inference and let the grammar emerge under single-bit supervision.

Our contributions are:
\begin{itemize}
    \item We introduce of a novel neural model tailored specifically to the learning of regular grammars from positive and negative acceptor examples.
    There are no limits on the complexity of the regular grammars it can learn.
    \item We describe in detail a procedure for the recovery of the learned grammars and parse trees from the internals of our model, warranting explainability.
    \item We systematically evaluate our model across the dimensions of grammar size, grammar complexity, and training data quantity.
\end{itemize}

\section{Related Work}
\label{section:related_work}
From the perspective of algorithmic learning, we divide the literature on learning of regular languages into two groups: one concerning the learning of finite automata and the associated regular expressions, the other addressing the problem of grammar induction.

\textit{Learning finite automata.} Early work on the subject leveraged Hidden Markov Models \cite{rabiner1986introduction} and probabilistic finite state machines \cite{carrasco1994learning} as the models for learning.
Later approaches readily recognized the utility of regular expressions as a form for description and tended to be deterministic.
Polynomial-time algorithms were proposed for learning of regular expressions without union operation from chosen classes of positive examples \cite{brazma1993efficient} and for learning of unambiguous regular expressions of maximum loop depth 2 \cite{kinber2010learning}. A genetic programming approach leveraging both positive and negative examples was used in \cite{svingen1998learning}, and a further polynomial-time method for $1$-unambiguous regular expressions aiming for simplicity of the resulting expressions was presented in \cite{fernau2009algorithms}.
Most of the recent work on regular expression learning takes the line of natural language processing (NLP) and proposes methods for particular uses in real-world datasets.
A genetic programming method for text extraction from XML documents is outlined in \cite{bartoli2014automatic}.
\cite{prasse2012learning} focuses on identifying email campaigns with high precision while phrasing it as an optimisation problem, and \cite{bui2014learning} gives a method based on Support-Vector Machines tailored to clinical texts.

Note that all of the above work either fences out a particular sub-class of regular expressions that are to be learned (e.g. union-less or $1$-unambiguous) or makes additional assumptions based on the nature of the particular real-world problem it is designed to solve.

\textit{Grammar induction.}
Pioneering work on the learning of grammars from examples often attempted to construct probabilistic context-free grammars that generated the target language $\mathcal{L}$ \cite{fu1975grammatical}.
Subsequent attempts employed a wider variety of tactics including Bayesian methods \cite{chen1995bayesian} and genetic programming \cite{wyard1993context}.
The learning was done chiefly from positive examples, though methods for automatic negative example generation from positive examples were later introduced \cite{smith2005guiding}.
Similarly to above, contemporary work on grammar induction is almost exclusively guided by the desired applications in NLP rather than interest in theory of computation.
As such, black-box recurrent neural automata \cite{choi2018learning}, reinforcement learning models \cite{yogatama2016learning}, and partially interpretable structured attention \cite{kim2017structured} and transformer \cite{goertzel2020guiding} models dominate the sub-field at present.

Our approach spans both of these groups.
We train on accept/reject information just like an acceptor automaton but learn a grammar.
In contrast to research on learning finite automata and regular expressions, there are no inherent limits on the complexity of grammars our model can learn -- it can learn any regular grammar, not a sub-class characterised by a particular type of regular expressions.
Further, in contrast with the recent efforts in grammar induction, the neural architecture we present learns a grammar in a fully explainable fashion.
Explainability is present in the training procedure, grammar extraction, and the use of our model for direct parsing by the learned grammar.

Given the above, our approach is more akin to research in program synthesis and fits better under the paradigm of ``algorithm learning''.
This is because we can view the language examples together with the annotations that mark them as positive and negative as input-output pairs for the broader \textit{programming by example (PbE)} task \cite{smith2000programming}.
In PbE, a program in a given programming language is to be synthesised based on a set of input-output pairs.
In our case, the programming language is the rewriting scheme for regular grammars and the program is the particular regular grammar that generates $\mathcal{L}$ or a substantial overlap therewith.

\section{Model}
\label{section:architecture}
Our network consists of three components:  \textit{Terminal Grammar Unit}, \textit{Non-terminal Grammar Unit}, and  \textit{Start Selector Unit}.
These are engaged sequentially following a simplified CYK algorithm to perform the function of an acceptor for the target language $\mathcal{L}$.
After training, the internals of each unit can be directly inspected to recover the learned regular grammar $\mathcal{G}' = \left(\mathcal{A}_T, \mathcal{A}_N', \mathcal{P}'\right)$.
Note that while the terminal alphabet $\mathcal{A}_T$ is considered known (from $\mathcal{L}, \mathcal{E}$) and shared with our model, the non-terminal alphabet $\mathcal{A}_N'$ is learned indirectly through use in learned productions $\mathcal{P}'$.
We present our method as for the learning of left-regular grammars, but the entire setup can be inverted to produce right-regular grammars.

\begin{figure*}[t!]
\centering
  \includegraphics[width=0.95\textwidth]{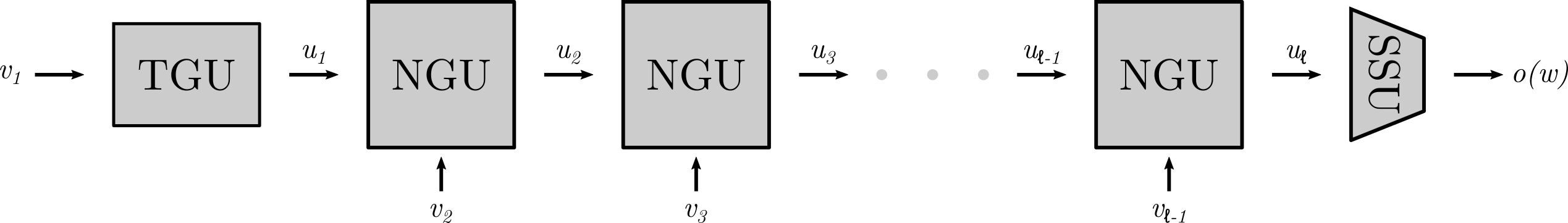}
  \caption{
    An overview of the model's neural structure.
    \textit{From left.} The Terminal Grammar Unit is applied to the first letter of the input word $w$.
    The non-terminal grammar unit is then applied recurrently (i.e. the trained weights are shared between individual instances) to the remaining letters.
    The final parse belief vector $u_{\ell}$ is fed into the Start Selector Unit to yield the accept/reject verdict.
  }
  \label{figure:blocks}
\end{figure*}

\subsection{Application Algorithm}
\label{section:application_algorithm}
For each example word $w{:}a_1a_2a_3\dots a_{\ell}$ we one-hot encode letters $a_i$ into $t$-dimensional vectors $v_i$, where $t = \left|\mathcal{A}_T\right|$ is the size of the terminal alphabet.

If $w$ is a positive example we associate it with label $y=1$, otherwise $y=0$.
$v_1$ is then fed into the Terminal Grammar Unit to produce an intermediate parse $n'$-dimensional (belief) vector $u_1$.
$n'$ is a parameter of the training, giving an upper bound on the number of non-terminals that may appear in $\mathcal{P}'$.
The ground-truth non-terminal alphabet $\mathcal{A}_N$ is not known to us during training and so neither is its size $n = \left|\mathcal{A}_N\right|$.
$n'$ is therefore just a guess for $n$, made in a thought process similar to that for the number of latent dimensions to be used in a disentangling variational autoencoder.

For $1 \leq i < \ell$, $v_{i+1}$ and $u_i$ are then provided as the input for the Non-terminal Grammar Unit, which in turn produces the intermediate parse vector $u_{i+1}$.

\begin{figure*}[t!]
\centering
  \includegraphics[width=0.90\textwidth]{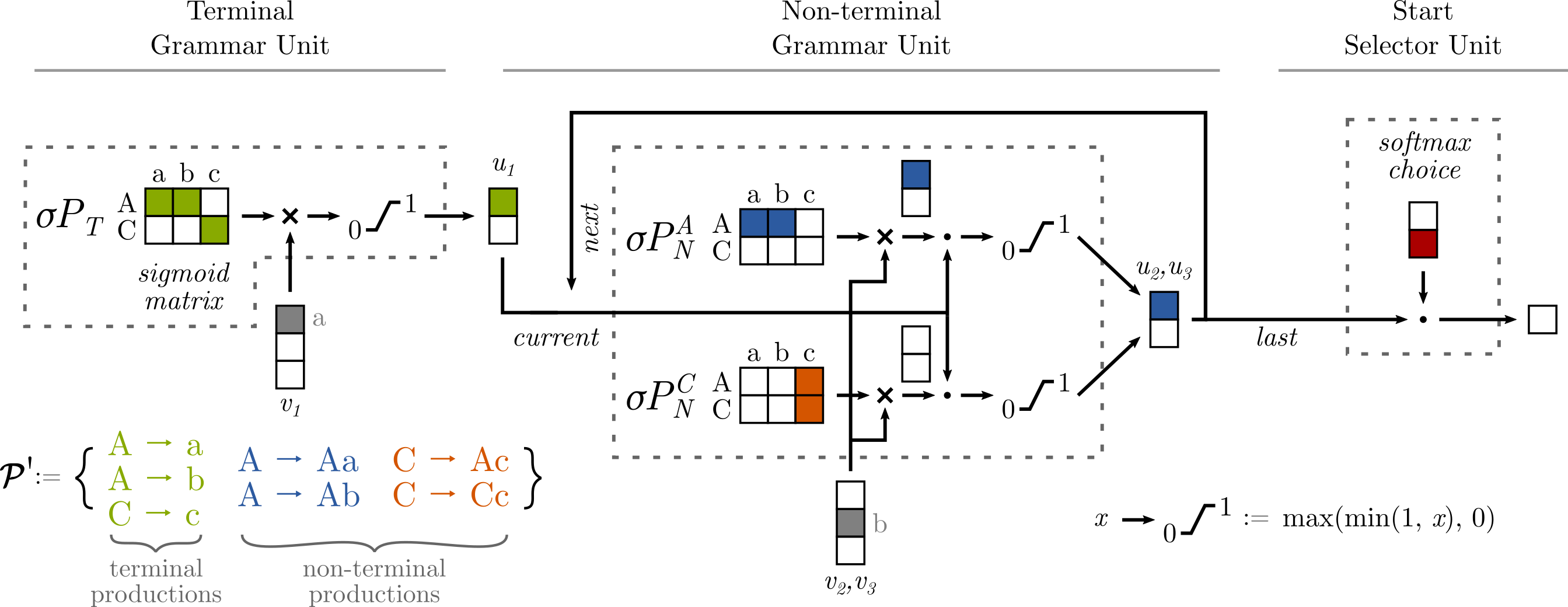}
  \caption{
    An illustration of the model's internals.
    The productions $\mathcal{P}'$ of the model hypothesis grammar $\mathcal{G}' = \left(\mathcal{A}_T, \mathcal{A}_N', \mathcal{P}'\right)$ generate the language given by the regular expression $\left(a|b\right)^*cc^*$.
    The $\times$ symbol represents matrix multiplication, $\cdot$ represents vector inner product, and the colour coding of productions and matrix entries marks equivalences by the encoding-decoding procedures of \Cref{section:ngu} and \Cref{section:grammar_extraction}.
    We present the model with a string $abb$.
    \textit{Left.} The Terminal Grammar Unit matches the initial $a$ encoded as $v_1 = (1\,0\,0)^{\text{T}}$ with the production $A \to a$ and produces the belief vector $u_1 = (1\,0)^\text{T}$.
    \textit{Middle}. The Non-terminal Grammar Unit has its production matrices queried by $v_2 = (0\,1\,0)^{\text{T}}$ and finds a match in the production $A \to Ab$.
    This is then dotted with the prior belief $u_1$ to produce the next parse belief $u_{2}$.
    The same is repeated for $v_3,u_2 = v_2,u_1$ giving $u_\ell = u_3 = (1\,0)^\text{T}$.
    \textit{Right.} The model has been trained to recognize $C$ as the root symbol, but the terminal parse $u_\ell$ on the given negative example is $(1\,0)^\text{T}$, leading to $0$ as the output of the acceptor.
  }
  \label{figure:internals}
\end{figure*}

Finally, $u_{\ell}$ is processed by the Start Selector Unit, which outputs a number $o(w)$ between $0$ and $1$ representing the belief that $w \in \mathcal{L}$.
The whole procedure is illustrated in \Cref{figure:blocks} on the unit level, and in \Cref{figure:internals} with internals and on a simple example.
In training, $o(w)$ is assigned a loss value computed as the binary cross-entropy between $y$ (true label) and $o(w)$ (predicted label).
The model is trained to minimise total loss combining the cross-entropy and two other losses reflecting the quality of the hypothesis grammar.

\subsection{Terminal Grammar Unit (TGU)}
The TGU serves as the hypothesis grammar parser for potential terminal productions.
It takes a $t$-dimensional vector $v_1$ -- the one-hot encoding of the first letter -- as input, uses it to query the trained hypothesis terminal production $n'$-by-$t$ matrix $P_T$, and clamps the output between $0$ and $1$:
$
    \text{TGU}\left( v_1 \right) := \text{clamp}\left(\sigma\left(P_T\right) v_1\right),
$
where $\text{clamp}(v) := \max\left(\min\left(1, v\right), 0\right)$ and $\sigma$ is the logistic sigmoid, each applied element-wise.
If $A$ and $b$ are one-hot-encoded as the $i$-th and $j$-th euclidean basis vectors, then the $i,j$-th entry of $\sigma\left(P_T\right)$ represents the belief of the model (the strength of TGU's hypothesis) that the production $A \to b$ is a part of the grammar $\mathcal{G}'$ being learned from $\mathcal{E}$.

\subsection{Non-terminal Grammar Unit (NGU)}
\label{section:ngu}
The NGU is the recurrent unit of our architecture and is the parser for the hypothesis grammar's non-terminal productions.
It holds $n'$ $n'$-by-$t$ trained matrices, each representing the hypothesised non-terminal productions with one of the $n'$ potential non-terminals $\mathcal{A}_N'$ on the left-hand side.
To perform its parse, the NGU takes the previous parse vector $u_i$, current terminal vector $v_{i+1}$, and yields $u_{i+1}$ where the $k$-th entry of $u_{i+1}$ for $1 \leq k \leq n'$ is given by
$
    \text{NGU}\left(u_i, v_{i+1} \right)_k := \text{clamp}\left(\left(\sigma\left(P_N^k\right) v_{i+1}\right)^{\text{T}}u_i\right).
$
If $A,B$ and $c$ are one-hot-encoded as the $k$-th, $i$-th, and $j$-th euclidean basis vectors respectively, then the $i,j$-th entry of $\sigma\left(P_N^k\right)$ represents the strength of NGU's hypothesis that the production $A \to Bc$ belongs to $\mathcal{P}'$.

\subsection{Start Selector Unit (SSU)}
The final parsing belief vector $u_\ell$ describes the model confidence about each of the $n'$ non-terminals being the root non-terminal.
In order for a parse to be successful, the parsing must terminate by reaching a particular non-terminal letter having the role of the start symbol.
We allow for the start symbol to emerge in an unsupervised fashion by letting the model learn which of the non-terminals should be considered to have the function of the root of $\mathcal{G}'$.
We achieved this by making a constant softmax choice (paying ``constant attention'') in the Start Selector Unit:
$
    o(w), \text{SSU}\left( u_\ell \right) := \text{softmax}\left(s\right)^\text{T} u_\ell,
$
where $s$ is a trained $n'$-dimensional vector.
We also experimented with using a two- and three-layer multi-layer perceptron (MLP) networks and found no difference in performance but observed a tendency of the model to encode the productions of the grammar's start symbol in the MLP, hindering explainability.
We also considered fixing one of the entries of $u_\ell$ but observed a decrease in recall scores.

\subsection{The Role of clamp$\left( \cdot \right)$}
If the hypothesis grammar is or nears being ambiguous, the vector inner product in NGU often results in belief values in $u_{i+1}$ being greater than $1$ (e.g. as in \Cref{figure:tree}).
To keep the model trainable and explainable at the same time, we limit all grammar unit outputs to $1$.
The values in our model never go below $0$, but we keep the lower bound in the definition of clamp$\left( \cdot \right)$ for consistency with similar work in neural networks.

\subsection{Training Losses}
\label{section:losses}
The training loss for our model consists of three components, with relative contributions to the total loss being controlled by hyperparameter factors.

\subsubsection{Prediction loss} The binary cross-entropy between the true labels ($1,0$ for positive,negative examples $w$) and predicted labels $o(w)$.
The prediction loss guides the model towards learning a grammar that generates exactly $\mathcal{L}$.
    
\subsubsection{Sharpening loss} For every entry $e$ in $\sigma\left(P_T\right),\sigma\left(P_N^k\right)$ we compute the sharpening contribution $
    \mathcal{S}\left( e \right) := 1 - \left(2e - 1\right)^2
$ and then compute the total sharpening loss $
    \mathcal{S} := \frac{1}{n'(n' + 1)t}\sum_e \mathcal{S}\left( e \right).
$
The sharpening loss helps to ensure that the values of $\sigma\left(P_T\right),\sigma\left(P_N^k\right)$ are eventually clearly interpretable as productions of $\mathcal{G}'$.
    
\subsubsection{Production use loss} Simply the mean of all entries $e$ of $\sigma\left(P_T\right),\sigma\left(P_N^k\right)$, i.e.
$
    \mathcal{U} := \frac{1}{n'(n' + 1)t} \sum_e e \,.
$
Intuitively, this loss encourages the use of smaller grammars in contrast to larger ones.

\subsubsection{Total loss}
The total loss for a batch $\mathcal{B}$ is then
$
    \ell(\mathcal{B}) := \text{BCE}\left(\mathcal{B}\right) + \beta \mathcal{S} + \gamma \mathcal{U},
$
where $\beta, \gamma \geq 0$ are hyperparameters.

\begin{figure}[t!]
\centering
  \includegraphics[width=0.45\textwidth]{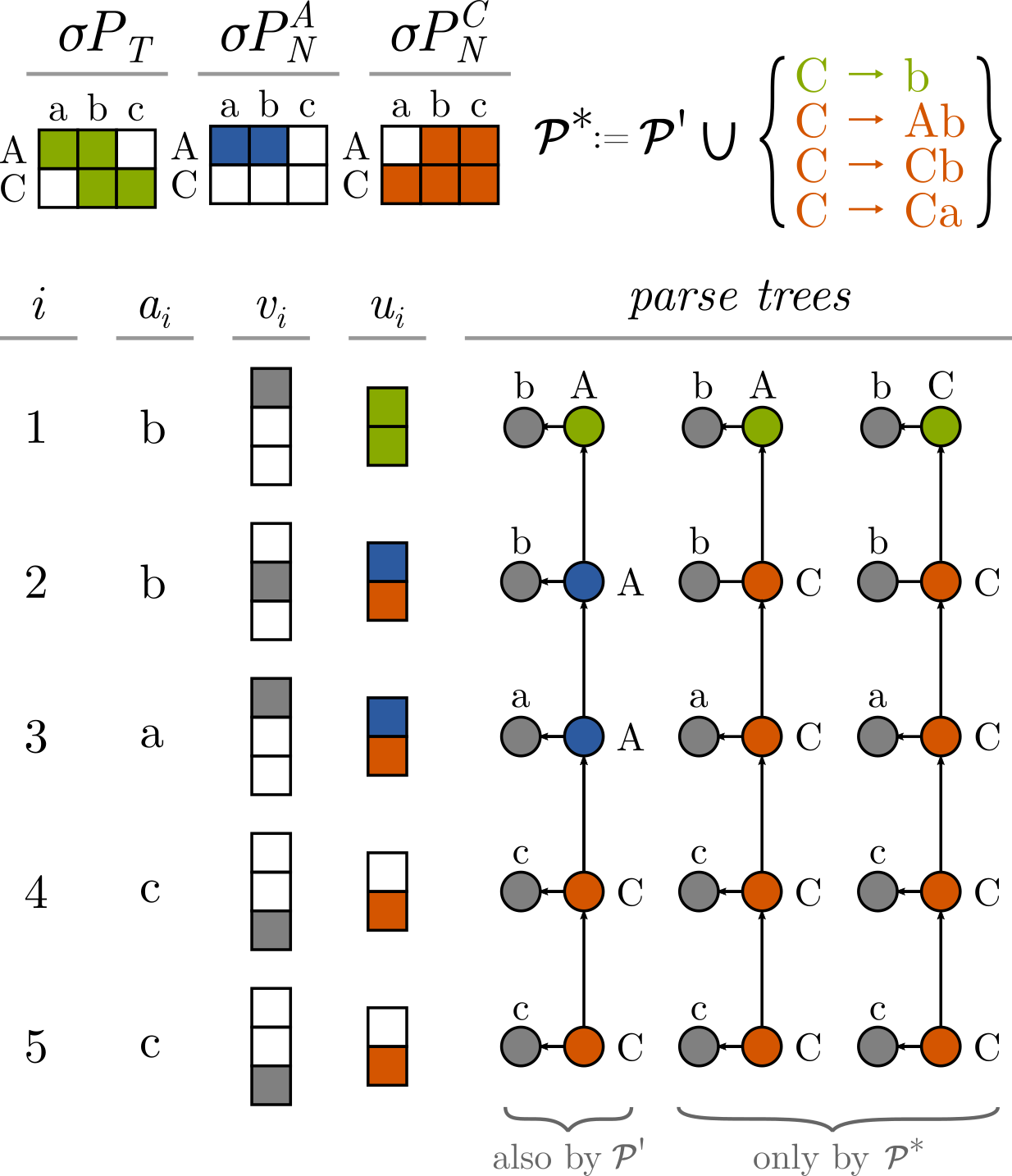}
  \caption{
    An example of parse tree construction from parsing input vectors $v_i,u_i$.
    Expanding on example from \Cref{figure:internals} we consider an instance of our model employing grammar $\mathcal{G}^* := \left(\mathcal{A}_T, \mathcal{A}_N', \mathcal{P}^*\right)$ to parse the word $bbacc$.
    $\mathcal{P}^*$ is an extension of $\mathcal{P}'$, with the additional productions introducing ambiguity leading to two more potential parse trees.
    The ambiguity leads to values exceeding $1$ in the dot product in the NGU and clamping is engaged to keep the values between $0$ and $1$.
    \textit{Table from the top.} The initial terminal $b$ can be parsed by either of $A$ or $C$.
    For $i=2$, each of $A \to Ab, C \to Ab, C \to Cb$ can be engaged producing partial parses rooted at $A$ or $C$.
    For $i=3$, either $A \to Aa$ or $C \to Ca$ can be used, conditional on the root of the previous intermediate parse.
    For $i=4,5$, $C \to Cc$ applies.
   }
  \label{figure:tree}
\end{figure}

\subsection{Grammar Extraction}
\label{section:grammar_extraction}
We set a confidence threshold $\tau$ for when an entry of a production matrix $P_\bullet$ is to be interpreted as signifying the presence of the production in the grammar.
In our experimentation, $\tau = 0.95$ proved to be a reliable choice, though any threshold strictly below $1$ is eventually achievable owing to the sharpening loss.

Denote $\left(M\right)_{ij}$ the $i,j$-th entry of a matrix $M$, $a_k$ the $k$-th terminal in $\mathcal{A}_T$, and $A_k$ the $k$-th non-terminal in $\mathcal{A}_N'$.
Then the productions $\mathcal{P}'$ of the induced grammar $\mathcal{G}'$ can be extracted from the TGU and NGU by the following procedure:
\begin{itemize}
    \item If $\left(\sigma(P_T)\right)_{ij} \geq \tau$, add $A_i \to t_j$ to $\mathcal{P}'$.
    \item For each $1 \leq k \leq n$, if $\left(\sigma(P_N^k)\right)_{ij} \geq \tau$ then add $A_k \to A_i a_j$ to $\mathcal{P}'$.
    \item Let $\mu := \argmax_k \text{softmax}\left( s \right)_k$. Add $S \to A_\mu$ to $\mathcal{P}'$.
\end{itemize}
Note that by constructing $\mathcal{P'}$ in this manner, $\mathcal{A}_N'$ may contain non-terminals that are never used or that can never be reached from the start symbol $S$ in a derivation.

\subsection{Parse Tree Construction}
While the extracted induced grammar forms a basis for parser construction, the TGU and NGU can be used to construct the parse trees directly.
For all $1 \leq i \leq \ell$, $u_\ell$ multi-hot-encodes all non-terminals that can be used to generate the prefix sub-word $w_{\leq i} = a_1 \dots a_i$ of $w$.
This can be done inductively in $i$ as follows: for $1 \leq j \leq n$, if $\left(\sigma(P^k) v_i\right)_j \geq \tau$ and $u_i \geq \tau$, $N_k$ is a parse tree node for $u_{i+1}$ with one child being the root of the parse tree for $u_i$ and the other being the leaf terminal $a_i$.
This is illustrated in \Cref{figure:tree}.
Such a tree may exist for every $j$, giving all the at most $n'$ possible parse tree roots for any prefix sub-word $w_{\leq i}$.


\section{Experiments}
\label{section:experiments}
To systematically evaluate the grammar induction performance of our model, we generate left-regular grammars of varying complexity using a randomized procedure, and then use the said grammars to produce training datasets consisting of positive and negative examples.
Once trained, the instances of our model are inspected for their induced grammars $\mathcal{G}'$ and tested for how closely they match the ground-truth $\mathcal{G}$.

Due to the lack of related work addressing the problem of general regular grammar induction from positive and negative examples (cf. \Cref{section:related_work}), the objective of our experimentation is to investigate the robustness of our method to increases in grammar complexity and potential brevity in training examples (i.e. the cases when the training examples may be plenty but are not long enough to be wholly confident about the recurrence in a production).
This is in line with the example evaluation approach taken in other work \cite{svingen1998learning,bartoli2014automatic,prasse2012learning}.

\subsection{Grammar Generation}
During the ground-truth grammar generation phase, we vary the number of terminals $t$, the number of non-terminals $n$, and the average number of productions per non-terminal $\mathfrak{p}$, thus controlling the complexity of the generated grammar.
For each non-terminal $A_i$, the number of productions is subsequently sampled from a geometric distribution with parameter $\pi = \mathfrak{p}^{-1}$.
Each production for $A_i$ is either a terminal production of the form $A_i \to a$ with probability $p_T = 0.4$ or a non-terminal production of the form $A_i \to A_j a$ with probability $p_N = 1 - p_T$.
$A_j$ is chosen uniformly at random from the set $\mathcal{A}_N$ of candidate non-terminals and $a$ is chosen uniformly at random from the set $\mathcal{A}_T$ of terminals.
We designate one symbol in $\mathcal{A}_N$ as the start symbol $S$ without preference.

In order for the language $\mathcal{L} = \left<\mathcal{G}\right>$ to be non-empty, we add a randomly sampled terminal production to the set $\mathcal{P}$ of productions if there are none, and furthermore ensure the derivation reachability of all non-terminals in $\mathcal{A}_N$ from the start non-terminal $S$ by adding a single production $A_i \to A_j a$ for each unreachable non-terminal $A_j$ and some reachable non-terminal $A_i$. $A_i,a$ are again taken at random.

\begin{figure}[t!]
\centering
  \includegraphics[width=0.49\textwidth]{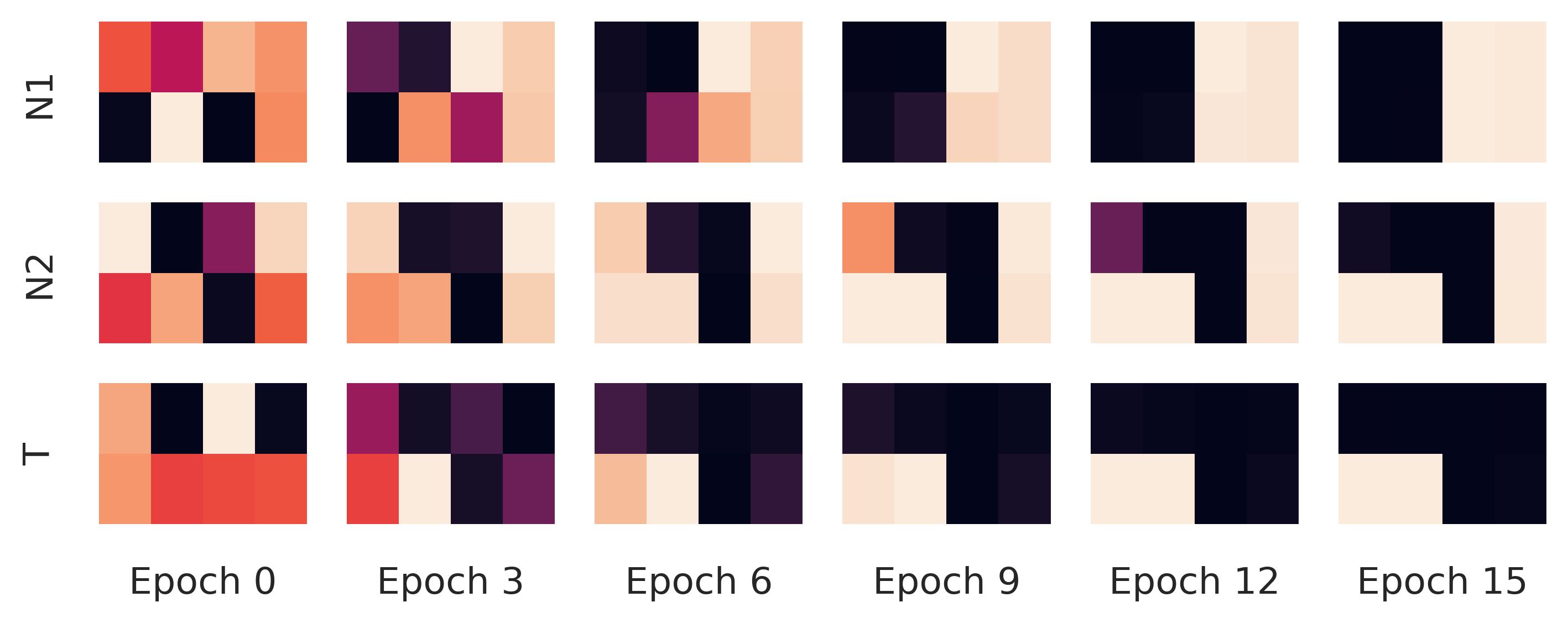}
  \caption{
    A visualisation of the training progression of a single run from the experiment described in \Cref{section:experiments}.
    The bright and dark tiles represent hypothesis grammar values close to $1$ and $0$ respectively.
    From left, we see random hypothesis grammar space becoming more orderly as the training progresses.
    The final induced grammar after the 15th epoch can be recovered following \Cref{section:grammar_extraction}.
  }
  \label{figure:progression}
\end{figure}

\subsection{Generation of Training Examples}
\label{section:generation_of_examples}
Let $\mathcal{G}$ be a ground-truth grammar generated as above, and let $\mathcal{L} = \left<\mathcal{G}\right>$ be its corresponding language.
The training data for a single instance of our model is a dataset $\mathcal{E}$ of words formed from the terminal alphabet $\mathcal{A}_T$, consisting of both negative and positive examples for $\mathcal{L}$.
To generate the examples for $\mathcal{L}$, we construct a minimal deterministic finite automaton (DFA) $\mathcal{D}$ equivalent to $\mathcal{G}$ in the sense that $\mathcal{D}$ accepts a word $w$ over $\mathcal{A}_T$ if and only if $w \in \mathcal{L}$.

Constructing finite automata to generate $\mathcal{E}$ instead of using the ground-truth $\mathcal{G}$ directly helps us avoid introducing unintentional bias into $\mathcal{E}$ that would favour $\mathcal{G}$ or grammars very similar to $\mathcal{G}$.
In other words, we avoid information leakage by constructing equivalent automata and optimising them prior to generating $\mathcal{E}$.

\subsubsection{Positive examples}
We perform a breadth-first search of depth $d$ on $\mathcal{D}$, memorising the path taken on each branch of the search.
Whenever an accepting state is encountered, the word consisting of  transition symbols of the given path read out in sequence is returned.
The search then continues to explore the path as before until the depth $d$ has been reached.
Observe that the length $\ell$ of the word is $\leq d$.

\subsubsection{Negative examples from non-accepting paths}
Paths (including intermediate paths, i.e. paths of length $\leq d$) of the above breadth-first search that do not end at an accepting state are added to $\mathcal{E}$ as negative examples.

\subsubsection{Negative examples with invalid postfix}
Let $\sigma$ be a state of $\mathcal{D}$ reached by the above search after $k$ steps.
Let $\mathcal{A}\left(\sigma\right)$ be the set of letters $a \in \mathcal{A}_T$ that label out-transitions of $\sigma$.
For all $a \in \mathcal{A}_T \backslash \mathcal{A}\left(\sigma\right)$, let $w$ be the word formed by appending $a$ to the $a_1\dots a_k$ labels of the transitions on the path traversed to reach $\sigma$.
Such $w$ are also negative examples.

\subsubsection{Negative examples with invalid infix}
Let $a_1\dots a_k a$ be a word for an invalid postfix negative example, and let $b_1\dots b_l$ be the symbols of some valid path between two states of $\mathcal{D}$ such that the latter state is accepting.
We consider any word formed as $a_1\dots a_k a b_1 \dots b_l$ a negative example.

\subsubsection{Random negative examples}
To further enrich $\mathcal{E}$, we generated a number of words of uniformly random length up to $d$ with each letter drawn from $\mathcal{A}_T$ uniformly at random such that $\mathcal{D}$ would not accept them, and added them to $\mathcal{E}$ as negatives.

Intuitively, the positive examples guide the model towards learning a grammar $\mathcal{G}'$ that recalls all words of $\mathcal{L}$ (i.e. $\mathcal{L} \subseteq \left<\mathcal{G}'\right>$),
the negative examples with invalid prefix or infix increase precision (i.e. minimise $\mathcal{L}^c \cap \left<\mathcal{G}'\right>$),
and the negative examples form non-accepting paths further contribute to increases in precision by discouraging spurious links between the hypothesis non-terminals $\mathcal{A}_N'$ of $\mathcal{G}'$ while also indirectly reducing redundancy of computational logic in $\mathcal{G}'$.

We experimented with various ratios of positive to negative examples and ended up settling for $1{:}1$, with random negative examples providing more negative examples wherever needed to reach this ratio.

\subsection{Evaluation}
\label{section:evaluation}
Given a trained model, we extract the induced grammar $\mathcal{G}'$ as per \Cref{section:grammar_extraction}.
We then convert $\mathcal{G}'$ into an equivalent DFA, from whom we compute the \emph{minimal} DFA  $\mathcal{D}'$.
The minimality is in the number of states as arrived at by the Hopcroft's algorithm.

For each regular language $\mathcal{L}$, there exists a unique (up to a re-labelling isomorphism) minimal recognizer DFA $\mathcal{D}$ \cite[p.~159-164]{hopcroft2001introduction}.
Given the minimal DFA $\mathcal{D}$ of the ground-truth grammar and the minimal DFA $\mathcal{D}'$ of the induced grammar, we canonically re-label their states and follow the algorithm of \cite{almeida2009testing} to test whether $\mathcal{D},\mathcal{D}'$ are isomorphic.
Isomorphism of $\mathcal{D},\mathcal{D}'$ means that $\mathcal{G},\mathcal{G}'$ are fully equivalent, implying that our model has achieved the maximum recall and precision and that we do not need to evaluate further.

In the case that $\mathcal{G},\mathcal{G}'$ are not equivalent, we assess similarity of languages by comparing the finite subsets $\mathcal{L}_d,\mathcal{L}_d'$ of each.
These consist of all words of $\mathcal{L}, \mathcal{L}' = \left<\mathcal{G}\right>$ up to length $d$.
We then measure the recall $\frac{\mathcal{L}_d \cap \mathcal{L}_d'}{\mathcal{L}_d}$, precision $\frac{\mathcal{L}_d \cap \mathcal{L}_d'}{\mathcal{L}_d'}$, and accuracy $\frac{\mathcal{L}_d \cap \mathcal{L}_d'}{\mathcal{L}_d \cup \mathcal{L}_d'}$ of our model.

\subsection{Results}
\label{section:results}
We generated grammars with $t = 4$, $n \in \left\{ 2, 3, 4 \right\}$, and $\mathfrak{p} \in \left\{2, 3, 4, 5 \right\}$, resulting in 12 different grammar configurations.
For each grammar, we generated the full dataset of positive and negative examples of strings up to length 16, training models on strings of length up to 6, 8, 10, 12, 14, and 16, with 5 runs per grammar per length.
During training, we used a batch size of 80, Adam optimizer with learning rate $0.005$, $\beta$ of $0.05$ after 60\% of epochs and $0$ before then.
There was a maximum of 60 epochs and $n' = 5$.
We extracted the grammar from the model using a confidence threshold $\tau = 0.95$, and evaluated it as described above.


Overall, the model learned the grammar from which the data was generated \textit{exactly} (i.e. the automata $\mathcal{D},\mathcal{D}'$ were isomorphic) in 85\% of all runs.

For more complex grammars, we observed an increase in accuracy (+2-5\%) when increasing length of examples while keeping the grammar complexity fixed.
Generally, grammars  with smaller numbers of non-terminals $n$ were easier to learn for our model, and for all $n$, higher number of productions per non-terminals also led to better results.
This is somewhat counter-intuitive but perhaps due to our methodology for negative example generation: more complex grammars had more negative examples from non-accepting paths and fewer random negative examples.

As pointed out in \Cref{section:related_work}, no previous method addressed the problem of learning the entire class of regular grammars from positive and negative examples, and so no direct comparison can be made at present.

\section{Conclusion}
\label{section:conclusion}
We have introduced a purely neural explainable model for the induction of regular grammars from positive and negative acceptor examples, and demonstrated its ability to induce grammars to high levels of accuracy.

Our model can be used both for grammar induction and as a regular language parser.
This duality of purpose arises from it simultaneously inducing a grammar and attempting partial parses throughout its training as an acceptor automaton.

The ultimate but distant goal of grammatical inference in the context of algorithm learning is the induction of general grammars (\textsc{Type}-0 in the Chomsky hierarchy) which are equivalent in power to Turing machines.
We see regular grammars as a simple but rich class of grammars for induction from examples under the Programming by Example paradigm and hope that our work will help to facilitate further advancements in explainable neural grammar inference.

\bibliographystyle{plain} 
\bibliography{bibliography} 

\begin{thebibliography}{10}

\bibitem{almeida2009testing}
Marco Almeida, Nelma Moreira, and Rog{\'e}rio Reis.
\newblock Testing the equivalence of regular languages.
\newblock {\em arXiv preprint arXiv:0907.5058}, 2009.

\bibitem{bartoli2014automatic}
Alberto Bartoli, Giorgio Davanzo, Andrea De~Lorenzo, Eric Medvet, and Enrico
  Sorio.
\newblock Automatic synthesis of regular expressions from examples.
\newblock {\em Computer}, 47(12):72--80, 2014.

\bibitem{brazma1993efficient}
Alvis Br{\=a}zma.
\newblock Efficient identification of regular expressions from representative
  examples.
\newblock In {\em Proceedings of the sixth annual conference on Computational
  learning theory}, pages 236--242, 1993.

\bibitem{bui2014learning}
Duy Duc~An Bui and Qing Zeng-Treitler.
\newblock Learning regular expressions for clinical text classification.
\newblock {\em Journal of the American Medical Informatics Association},
  21(5):850--857, 2014.

\bibitem{carrasco1994learning}
Rafael~C Carrasco and Jose Oncina.
\newblock Learning stochastic regular grammars by means of a state merging
  method.
\newblock In {\em International Colloquium on Grammatical Inference}, pages
  139--152. Springer, 1994.

\bibitem{chen1995bayesian}
Stanley~F Chen.
\newblock Bayesian grammar induction for language modeling.
\newblock {\em arXiv preprint cmp-lg/9504034}, 1995.

\bibitem{choi2018learning}
Jihun Choi, Kang~Min Yoo, and Sang-goo Lee.
\newblock Learning to compose task-specific tree structures.
\newblock In {\em Proceedings of the AAAI Conference on Artificial
  Intelligence}, volume~32, 2018.

\bibitem{fernau2009algorithms}
Henning Fernau.
\newblock Algorithms for learning regular expressions from positive data.
\newblock {\em Information and Computation}, 207(4):521--541, 2009.

\bibitem{fu1975grammatical}
King-Sun Fu and Taylor~L Booth.
\newblock Grammatical inference: Introduction and survey-part ii.
\newblock {\em IEEE Transactions on Systems, Man, and Cybernetics},
  (4):409--423, 1975.

\bibitem{goertzel2020guiding}
Ben Goertzel, Andr{\'e}s Su{\'a}rez-Madrigal, and Gino Yu.
\newblock Guiding symbolic natural language grammar induction via
  transformer-based sequence probabilities.
\newblock In {\em International Conference on Artificial General Intelligence},
  pages 153--163. Springer, 2020.

\bibitem{hopcroft2001introduction}
John~E Hopcroft, Rajeev Motwani, and Jeffrey~D Ullman.
\newblock Introduction to automata theory, languages, and computation.
\newblock {\em Acm Sigact News}, 32(1):60--65, 2001.

\bibitem{kim2017structured}
Yoon Kim, Carl Denton, Luong Hoang, and Alexander~M Rush.
\newblock Structured attention networks.
\newblock {\em arXiv preprint arXiv:1702.00887}, 2017.

\bibitem{kinber2010learning}
Efim Kinber.
\newblock Learning regular expressions from representative examples and
  membership queries.
\newblock In {\em International Colloquium on Grammatical Inference}, pages
  94--108. Springer, 2010.

\bibitem{prasse2012learning}
Paul Prasse, Christoph Sawade, Niels Landwehr, and Tobias Scheffer.
\newblock Learning to identify regular expressions that describe email
  campaigns.
\newblock {\em arXiv preprint arXiv:1206.4637}, 2012.

\bibitem{rabiner1986introduction}
Lawrence Rabiner and Biinghwang Juang.
\newblock An introduction to hidden markov models.
\newblock {\em ieee assp magazine}, 3(1):4--16, 1986.

\bibitem{smith2000programming}
David~Canfield Smith, Allen Cypher, and Larry Tesler.
\newblock Programming by example: novice programming comes of age.
\newblock {\em Communications of the ACM}, 43(3):75--81, 2000.

\bibitem{smith2005guiding}
Noah~A Smith and Jason Eisner.
\newblock Guiding unsupervised grammar induction using contrastive estimation.
\newblock In {\em Proc. of IJCAI Workshop on Grammatical Inference
  Applications}, pages 73--82, 2005.

\bibitem{svingen1998learning}
Borge Svingen.
\newblock Learning regular languages using genetic programming.
\newblock In {\em Proc. 3-rd Genetic Programming Conference}, pages 374--376,
  1998.

\bibitem{wyard1993context}
Peter Wyard.
\newblock Context free grammar induction using genetic algorithms.
\newblock In {\em IEE colloquium on grammatical inference: theory, applications
  and alternatives}, pages P11--1. IET, 1993.

\bibitem{yogatama2016learning}
Dani Yogatama, Phil Blunsom, Chris Dyer, Edward Grefenstette, and Wang Ling.
\newblock Learning to compose words into sentences with reinforcement learning.
\newblock {\em arXiv preprint arXiv:1611.09100}, 2016.

\end{thebibliography}

\end{document}